\documentclass[11pt]{article}

\usepackage{amsmath,amsbsy,amsfonts,amssymb,amsthm,units}
\usepackage{graphicx}
\usepackage{mathtools,appendix}
\usepackage{color,cases}
\usepackage{tikz}
\usetikzlibrary{calc,shapes}
\usepackage{subfigure}
\usepackage{natbib}
\usepackage{hyperref}
\usepackage{algorithm}
\usepackage{algorithmic}

\usepackage{color,cases}
\usepackage{appendix}
\usepackage{enumitem}

 \newcommand{\IGNORE}[1]{}

\def\nn{\nonumber}

\newcommand\E{\mathbb{E}}
\newcommand\R{\mathbb{R}}

\newcommand\inner[1]{\ensuremath{\langle #1 \rangle}}

\def\tha{{\mbox{\tiny th}}}

\DeclareMathOperator{\Diag}{Diag}

 \def\0{{\bf 0}}

\def\nn{\nonumber}

\def\qed{\hfill\hbox{${\vcenter{\vbox{
    \hrule height 0.4pt\hbox{\vrule width 0.4pt height 6pt
    \kern5pt\vrule width 0.4pt}\hrule height 0.4pt}}}$}}

\definecolor{myred}{rgb}{0.3,0.0,0.7}
\definecolor{dkg}{rgb}{0.1,0.7,0.2}
\definecolor{dkb}{rgb}{0.0,0.2,0.8}

 \def\ha{\hat{a}}

\def\Nc{{\cal N}}

\def\Sc{{\cal S}}

\def\Ebb{{\mathbb E}}

\def\Rbb{{\mathbb R}}

\newcommand{\bprf}{\begin{myproof}}
\newcommand{\eprf}{\end{myproof}}
\newcommand{\bp}{\begin{psfrags}}
\newcommand{\ep}{\end{psfrags}}
\newcommand{\bl}{\begin{lemma}}
\newcommand{\el}{\end{lemma}}
\newcommand{\bt}{\begin{theorem}}
\newcommand{\et}{\end{theorem}}
\newcommand{\bc}{\begin{center}}
\newcommand{\ec}{\end{center}}
\newcommand{\bi}{\begin{itemize}}
\newcommand{\ei}{\end{itemize}}
\newcommand{\ben}{\begin{enumerate}}
\newcommand{\een}{\end{enumerate}}
\newcommand{\bd}{\begin{definition}}
\newcommand{\ed}{\end{definition}}
\def\beq{\begin{equation}}
\def\eeq{\end{equation}\noindent}
\def\beqn{\begin{eqnarray}}
\def\eeqn{\end{eqnarray} \noindent}
\def\beqnn{  \begin{eqnarray*}}
\def\eeqnn{\end{eqnarray*}  \noindent}
\def\bcase{  \begin{numcases}}
\def\ecase{\end{numcases}   \noindent}
\def\bsbcase{  \begin{subnumcases}}
\def\esbcase{\end{subnumcases}   \noindent}

\newtheorem{theorem}{Theorem}

\newtheorem{lemma}[theorem]{Lemma}

\newtheorem{definition}{Definition}

\newenvironment{myproof}{\noindent{\bf Proof:} \hspace*{1em}}{
    \hspace*{\fill} $\Box$ }
\newenvironment{proof_of}[1]{\noindent {\bf Proof of #1: }}{\hspace*{\fill} $\Box$ }

\newcommand{\matplottc}[1]{               
        \unitlength .45truein
        \begin{center}
        \includegraphics{#1.ps}
        \end{picture}
        \end{center}
}

\def\psfancypar#1#2{\begingroup\def\par{\endgraf\endgroup\lineskiplimit=0pt}
               \setbox2=\hbox{\large\sc #2}
               \newdimen\tmpht \tmpht \ht2 \advance\tmpht by \baselineskip
               \font\hhuge=Times-Bold at \tmpht
               \setbox1=\hbox{{\hhuge #1}}
               \count7=\tmpht \count8=\ht1
               \divide\count8 by 1000 \divide\count7 by \count8
               \tmpht=.001\tmpht\multiply\tmpht by \count7
               \font\hhuge=Times-Bold at \tmpht
               \setbox1=\hbox{{\hhuge #1}}
               \noindent
                \hangindent1.05\wd1
               \hangafter=-2 {\hskip-\hangindent
               \lower1\ht1\hbox{\raise1.0\ht2\copy1}%
                \kern-0\wd1}\copy2\lineskiplimit=-1000pt}

\def\Kout{\setbox1=\hbox{\Huge\bf K}\hbox to
1.05\wd1{\hspace{.05\wd1}
}}
\def\Sout{\setbox1=\hbox{\Huge\bf S}\hbox to 1.05\wd1{\hspace{.05\wd1}
}}

\newcommand{\torestate}[3]{%
\expandafter \def \csname BBRESTATE #2 \endcsname{#3}
\theoremstyle{plain}
\newtheorem{BBRESTATETHMNUM#2}[theorem]{#1}
\begin{BBRESTATETHMNUM#2}\label{#2}\csname BBRESTATE #2 \endcsname   \end{BBRESTATETHMNUM#2}
\newtheorem*{BBRESTATETHMNONNUM#2}{{#1}~\ref{#2}}
}

\newcommand{\restate}[1]{\begin{BBRESTATETHMNONNUM#1}[Restated] \csname BBRESTATE #1 \endcsname
\end{BBRESTATETHMNONNUM#1}}

\definecolor{blue1}{HTML}{0066FF}
\definecolor{lpurple}{cmyk}{.05,0.18,0,0}

\def\Pc{{\mathcal{S}}}

\usepackage{fullpage}

\author{Hanie Sedghi\footnote{University of California, Irvine. Email: sedghih@uci.edu} \quad Majid Janzamin\footnote{University of California, Irvine. Email: mjanzami@uci.edu}\quad Anima Anandkumar\footnote{University of California, Irvine. Email: a.anandkumar@uci.edu}}

\date{}
\title{Provable Tensor Methods for Learning \\ Mixtures of Generalized Linear Models}

\begin{document}
\maketitle

\begin{abstract}We consider the problem of learning mixtures of generalized linear models (GLM) which arise in   classification and regression problems.   Typical learning  approaches such as expectation maximization (EM) or variational Bayes  can get stuck in spurious local optima. In contrast,  we present a tensor decomposition method which is guaranteed to correctly recover the parameters. The key insight is to employ certain feature transformations of the input, which depend on the input generative model.
Specifically, we employ score function tensors of the input  and compute their cross-correlation with the response variable.   We establish that the decomposition of this tensor consistently recovers the parameters, under mild non-degeneracy conditions.  We demonstrate that the computational and  sample complexity of our method is a low order polynomial of the input and the latent  dimensions.
\end{abstract}


\section{Introduction}

A generalized linear model (GLM) is a flexible extension  of linear regression which allows the response or the output to be a non-linear function of the input via an {\em activation} function. In other words, in a GLM,  the linear regression of the input is passed through an activation function to generate the response. GLMs unify popular frameworks such as logistic regression and Poisson regression with linear regression. At the same time, they can be learnt with guarantees  using simple iterative methods~\citep{kakade2011efficient}.

In many scenarios, however,   GLMs may be too simplistic, and mixtures of GLMs can be much more effective since they combine the expressive power of latent variables with the predictive capabilities of the GLM. Mixtures of GLMs have widespread applicability including  object recognition~\citep{quattoni2004conditional}, human action recognition~\citep{wang2009max}, syntactic parsing~\citep{petrov2007discriminative}, and machine translation~\citep{liang2006end}.

%


Traditionally,   mixture models are learnt through heuristics such as expectation maximization (EM)~\citep{jordan1994hierarchical,xu1995alternative} or variational Bayes~\citep{svensen2003bayesian}. However, these methods can converge to spurious local optima and have slow convergence rates for high dimensional models. In contrast,   we employ a method-of-moments approach for guaranteed learning of mixtures of GLMs.

The method of moments paradigm dates back to Pearson~\citep{Pearson94}, and involves fitting the observed moments to parametric distributions.  Recently, it has been highly successful in unsupervised learning of a wide range of latent variable models such as Gaussian mixtures, topic models, hidden Markov models~\citep{AnandkumarEtal:tensor12}, network community models~\citep{AnandkumarEtal:community12COLT},  mixture of ranking models~\citep{pranjal,seewong}, and so on. The basic idea is to find an efficient spectral decomposition of low order observed moment tensors. Under natural non-degeneracy assumptions, the tensor method is guaranteed to correctly recover the underlying model parameters with low computational and sample complexities. Moreover, in practice, these methods are embarrassingly parallel and scalable to large-scale datasets~\citep{AnandkumarEtal:communityimplementation13}.

Earlier works on tensor methods~\citep{AnandkumarEtal:tensor12} consider unsupervised learning and a key assumption is that the observables are linear functions of the latent variables (in expectation). However, here, we consider mixtures of GLMs, which are non-linear, and this   rules out a direct application of tensor methods.

We address the above challenges with the following insight:  we have additional flexibility in the regression  setting since we have both the response and the input. We can therefore form different moments involving transformations of the  input and the response. What are the appropriate transforms for forming the moments which are amenable to tensor decomposition methods?
As detailed below, the key ingredient is using a specific  feature transformation of the input, based on its probability distribution.

\subsection{Summary of Result}
The main contribution of this work is to provide a guaranteed method for learning mixtures of GLMs using score function transformations.  The $m^{\tha}$ order score function
$\Pc_m(x)$ is related to the normalized $m^{\tha}$ order derivative of the   pdf of input $x$, see~\eqref{eqn:high-score}. We assume knowledge of these score functions, and this can be estimated via various unsupervised learning methods using only unlabeled samples (e.g. spectral methods).

We then construct the cross-moment tensor between    the response variable  and  the input score function. We establish that the decomposition of this tensor consistently recovers the components of the GLM mixture under some simple non-degeneracy assumptions.
Let the response or the output $y$ be generated from a mixture of GLMs: $\Ebb[y|h,x]= g(\inner{Uh, x})+\inner{\tilde{b},h})$,  where $g(\cdot)$ is the activation function, $x$ is the input and $h$ is the hidden choice variable. Let  $r$ be the number of mixture components, $d$ be the input dimension and $s_{\min}(U)$ be the $r^{\tha}$ largest singular value of $U$.
 Assume the  weight matrix $U=[u_1|\ldots|u_r]$ is full column rank. Then, we have the following result.
\begin{theorem}[Informal Result]
 We  recover the weight vectors $\{u_i\}$ (up to scaling) by performing tensor decomposition on the cross-moment tensor $\Ebb[y \cdot \Pc_3(x)]$. If we have $n=\tilde{O}\left(\frac{d^3 r^4}{\epsilon^2 s_{\min}^2(U)}\right)$ samples, the error in recovering each weight vector $u_i$ is bounded by $\epsilon$.
\end{theorem}

The above result requires third order score function $\Pc_3(x)$ to consistently estimate the weight vectors $\{u_i\}$ of the GLM components in the mixture. Note that the second order score function $\Pc_2(x)$  is only a matrix (assuming a vector input $x$) and can only identify the weights $\{u_i\}$  up to the subspace. Thus, we require at least the third order score function to consistently estimate the GLM mixture model. When the number of components exceeds the input dimension, the full column rank assumption on $U$ is violated, and in this case, we can  resort to higher order score functions to consistently estimate the parameters.

We  employ the tensor decomposition methods from~\citep{AnandkumarEtal:tensor12,  JanzaminEtal:Altmin14} to learn  the weight vectors $u_i$ (up to scaling). The tensor method is efficient to implement and does not suffer from spurious local optima.
 Thus, we guarantee consistent estimation of the weight vectors of GLM mixtures   through  decomposition of the cross-moment tensor involving the response variable and the input score functions. Our method is shown in Algorithm~\ref{algo:main}.

\subsection{Overview of Techniques}

\paragraph{Representation learning is the key: }A crucial ingredient in this work  is to first  learn the probabilistic model of the input, and  employ transformations based on the model for learning the GLM mixture. Thus, we characterize how unsupervised learning on the input can be carried over for learning conditional models of the output via tensor methods.

The  feature transformations we employ     are the (higher order) score functions,\footnote{In this paper, we refer to the derivative of the log of the density function with respect to the variable as the score function. In other works, typically, the derivative is taken with respect to some model parameter~\citep{jaakkola1999exploiting}. Note that if the model parameter is a location parameter, the two quantities only differ in the sign. Higher order score functions involve higher order derivatives of the density function. For the exact form, refer to~\citep{janzamin2014matrix}.} which capture local variation of the probability density function of the input. This follows a recent key  result  that the cross-moments between the response variable and the input  score functions  yield (expected) derivatives of the response, as a function of the input~\citep{janzamin2014matrix}.

\paragraph{Incorporating score functions into tensor decomposition framework: }In this paper, we exploit the above   result to form the expected derivatives  of the output as a function of the input. We then show that the expected derivatives have a nice relationship with the unknown parameters of the GLM mixture, and the form reduces to a tensor CP decomposition form. We require only a mild assumption on the activation function that it has non-vanishing third derivative (in expectation). For linear regression, this condition is violated, but we can easily overcome this by considering higher powers of the output in the moment estimation framework.
 
%


\section{Problem Formulation}

\paragraph{Notations: }
Let $[n]$ denote the set $\{1,2,\dotsc,n\}$. Let $e_i \in \Rbb^d$ denote the standard basis vectors in $\Rbb^d$. Let $I_d\in \Rbb^{d\times d}$ denote the identity matrix. $\tilde{\mathcal{O}}$ denotes the order when ignoring polylog factors. Throughout this paper, $\nabla_x^{(m)}$ denotes the $m$-th order derivative w.r.t.\ variable $x$ and notation $\otimes$ represents tensor (outer) product.

A real \emph{$p$-th order tensor} $T \in \bigotimes_{i=1}^p \R^{d_i}$ is a member of the tensor product of Euclidean spaces $\R^{d_i}$, $i \in [p]$.
As is the case for vectors (where $p=1$) and matrices (where $p=2$), we may
identify a $p$-th order tensor with the $p$-way array of real numbers $[
T_{i_1,i_2,\dotsc,i_p} \colon i_1,i_2,\dotsc,i_p \in [d] ]$, where
$T_{i_1,i_2,\dotsc,i_p}$ is the $(i_1,i_2,\dotsc,i_p)$-th coordinate of $T$
with respect to a canonical basis.


\paragraph{CP decomposition and tensor rank:} A $3$rd order tensor $T \in \Rbb^{d \times d \times d}$ is said to be rank-$1$ if it can be written in the form
$T=  a \otimes b\otimes c \Leftrightarrow T(e_i,e_j,e_l) =   a(i) \cdot b(j) \cdot c(l),$
where notation $\otimes$  represents the {\em tensor product}.
A tensor $T  $ is said to have a CP rank $k\geq 1$ if it can be written as the sum of $k$ rank-$1$ tensors
$T = \sum_{i\in [k]}  a_i \otimes b_i \otimes c_i.$

\subsection{Learning Problem}


Let $y$ denote the output and $x \in \mathbb{R}^{d}$ be the input. We consider both the regression setting, where $y$ can be continuous or discrete, or the classification setting, where $y$ is discrete. For simplicity, we assume $y$ to be a scalar: in the classification setting, this corresponds to binary classification  ($y \in \lbrace-1,1 \rbrace$).

We consider the {\em realizable} setting, where we assume that the output $y$ is drawn from an associative model $p(y|x)$, given input $x$. In addition,  we assume that the input $x$ is drawn from some continuous probability distribution with density function $p(x)$. We will incorporate this generative model  in our algorithm for learning the associative model.

We first consider mixtures of generalized linear models (GLM)~\citep{agarwal2013least,kakade2011efficient} and then extend to mixture of GLMs with nonlinear transformations.
The class of GLMs  is given by
\beq \label{eqn:glm} \Ebb[y|x] =   g(\inner{u,x}+b), \eeq
  where $g$ is   the {\em  activation} function, $u$ is the weight vector, and $b$ is the bias. $g(\cdot)$ is usually chosen to be the logistic function, although we do not impose this limitation.  In the binary classification setting, \eqref{eqn:glm} corresponds to a single classifier. 
  Note that a linear regression can be modeled  using a linear activation function. Throughout this paper we assume that noise is independent of the input. 

\noindent A mixture of $r$ GLM models is then given by employing a hidden choice variable $h \in \{e_1, e_2, \ldots, e_r\}$, where $e_i$ is the basis vector in $\mathbb{R}^r$ to select of the $r$ GLM models, i.e. \beq\label{eqn:glm-mix}\Ebb[y|x,h] = g(\inner{Uh,x}+\inner{\tilde{b},h}), \eeq where $U=[u_1|u_2\ldots u_r] \in \mathbb{R}^{d_x \times r}$ has the $r$ weight vectors of component GLMs as columns and $\tilde{b} \in \mathbb{R}^r$ is the vector of biases for the component GLMs. Let $w:=\Ebb[h]$ be the probability vector for selecting the different GLMs.

We then extend our results to learning mixture of  GLMs  with nonlinear transformations where
\beq \Ebb[y|x,h] = g(\inner{Uh, \phi(x)}, \inner{\tilde{b}, h}), \label{eqn:nonlinearmix}\eeq for some known function $\phi(\cdot)$. 

Given training samples $\{x_i, y_i\}$, our goal is to learn the parameters of the associative mixture described above. We consider a moment-based approach, which involves cross-moments of $y$ and a function of $x$. We first assume that the exact moments are available, and we later carry out sample analysis, when empirical moments are used.

Throughout this paper we make the following assumptions unless otherwise stated.
Derivative and expectation are interchangeable.
The activation function $g$ is differentiable up to the third order. The choice variable is independent of the input $x$, i.e., $h$ does not depend on $x$. The score function $\nabla_x \log p(x)$ exists and all the entries of $g(x) \cdot p(x)$ go to zero on the boundaries of support of $p(x)$.

\section{Learning   under Gaussian Input} \label{section:results}

We now present the method for learning the mixture models in \eqref{eqn:glm-mix} and \eqref{eqn:nonlinearmix}. We first start with the simple case, where the input $x$ is Gaussian, and we have a single GLM model, instead of a mixture, and then extend to more general cases.

\begin{algorithm}[t]
\caption{Learning mixture of associative models $\Ebb[y|x,h] = g(\inner{Uh,x}+\inner{\tilde{b},h})$}
\label{algo:main}
\begin{algorithmic}[1]
\renewcommand{\algorithmicrequire}{\textbf{input}}
\renewcommand{\algorithmicensure}{\textbf{output}}
\REQUIRE Labeled samples $(x_i,y_i), i \in [n]$.
\REQUIRE Score function of the input $\Pc_3(x)$ as in Equation~\eqref{eqn:high-score}.

\STATE Compute $\widehat{M}_3=\frac{1}{n}\sum_i y_i \cdot  \Pc_3(x_i)$, Empirical estimate of $M_3$. 
\STATE $\lbrace \hat{u}_j \rbrace_{j \in [r]}=\text{tensor power decomposition}(\widehat{M}_3)$. (Algorithm~\ref{alg:robustpower} in the Appendix) \label{line:weight}
\STATE Recover scale and biases using EM (as in Appendix~\ref{sec:EM} ). 
\end{algorithmic}
\end{algorithm}

 \subsection{Toy Example: single GLM}

We first assume   a white Gaussian  input  $x \sim \mathcal{N}(0,I_{d})$ to demonstrate our ideas. Assuming that $y$ is generated from a GLM
\[ \Ebb[y|x] = g(\inner{u,x}+b),\] we have the following result on the cross-moment $\Ebb[y\cdot x]$.

 \begin{lemma}[Moment form for Gaussian input and single GLM] \label{lemma:binary}We have\begin{align*} M_1&=\Ebb[y\cdot x]=\Ebb[\nabla_{x'} g(x')]\cdot u,
 \end{align*}where the expectation is over $x':= \inner{u,x} + b$, and $ x\sim \Nc(0,I_{d})$.\end{lemma}

Proof follows from Stein's identity~\citep{stein1972} as discussed below. Thus, by forming the first-order cross-moment $M_1$, we can recover the weight vector $u$ up to scaling. Note that the scaling and the bias $b$ are just scalar parameters which can be estimated separately.

The main message behind Lemma~\ref{lemma:binary} is that the cross-moments between the output $y$ and the input $x$ contain valuable information about the associative model. In the special case of Gaussian input and single GLM, the first order moment is sufficient to learn almost all the parameters of the GLM. But how general is this framework? Can we use a moment-based framework when there are mixture of GLMs? We exhibit that higher order moments can be used to learn the GLM mixture under Gaussian input in the next section. What about the case when the input is not Gaussian, but has some general distribution? We consider this setting in Section~\ref{sec:general_distr} and show that surprisingly we can form the appropriate cross-moments for learning under  any general (continuous) input distribution.

%
\paragraph{Stein's Identity: }The proof of Lemma~\ref{lemma:binary} follows   from the Stein's identity for Gaussian distribution.  It states that  for all   functions $G(x)$ satisfying mild regularity conditions, we have~\citep{stein1972} 
\begin{equation} \label{eqn:Steinbasic}
\Ebb[G(x) \cdot x]= \Ebb[\nabla_x G(x)].
\end{equation}
Thus, Lemma~\ref{lemma:binary} is a direct application of the Stein's identity by substituting $G(x)$ with $g(\inner{u,x}+b)$.

\subsection{Learning GLM mixtures} \label{sec:gausstoy}

We now consider learning mixture of GLMs\[\Ebb[y|x,h] = g(\inner{Uh,x}+\inner{\tilde{b},h}), \] where $U=[u_1|u_2\ldots u_r]$ has the $r$ weight vectors of component GLMs as columns and $\tilde{b}$ is the vector of biases for the component GLMs. Recall that $w:=\Ebb[h]$ is the probability vector for selecting different GLMs.

For the mixture of GLMs, the first order moment $M_1:=\Ebb[y\cdot x]$ is now a combination of (scaled) weight vectors $u_i$'s, i.e.
\[M_1:= \Ebb[y\cdot x]=\sum_{j\in [r]} w_j \Ebb[\nabla_{x'_j}g(x'_j)] u_j, \] where the expectation is over $x'_j=\inner{u_j,x}+\tilde{b}_j$, and $x\sim \Nc(0,I_{d})$. Thus, the first order moment does not suffice for learning mixture of GLMs.

Now, let us look at the  second order  moment,
\[ M_2:=\Ebb[y\cdot (x\otimes x-I)]=\sum_{j\in [r]} \Ebb[\nabla_{x'_j}^{(2)}g(x'_j)] w_j \cdot u_j \otimes u_j ,\]where, as before, the expectation is over $x'_j=\inner{u_j,x}+\tilde{b}_j$. If the expectations (and $w_j$'s) are non-zero, then we can recover the subspace spanned by  the weight vectors $u_j$'s. However, we cannot recover the individual weight vectors $u_j$'s. Moreover, if the biases $\tilde{b}=0$ and $g$ is a symmetric function, then the expectations are zero, and the second order moment $M_2$ vanishes. A mirror trick is introduced in~\citep{sun2013learning} to alleviate this problem, but this still only recovers the subspace spanned by the $u_j$'s.
%

We now consider the third order moment $M_3$ in the hope of recovering the weight vectors $u_j$'s for mixture of GLMs.
We   show that by adjusting the moment $\Ebb[y\cdot x\otimes x\otimes x]$ appropriately, we obtain a CP tensor form in terms of the weight vectors $u_j$'s. Specifically, consider
 \begin{align}
 \label{eq:simple_third}
  M_3&:=\Ebb[y\cdot x\otimes x \otimes x]-\sum_{j \in [d]} \!\Ebb[y \cdot e_j \otimes x\otimes e_j] \\ \nonumber
  &~-\!\!\sum_{j \in [d]} \!\Ebb[y\cdot e_j \otimes e_j\otimes x] \!-\!\! \sum_{j \in [d]}\! \Ebb[y\cdot x \otimes e_j \otimes e_j].
  \end{align}

Note that $M_3$ can be considered as a special case of the form $\Ebb[y \cdot \Pc_3(x)]$ for white Gaussian  input  $x \sim \mathcal{N}(0,I_{d})$, where $\Pc_3(x)$ is the third order score function of the input as defined in Section~\ref{sec:stein}.

\begin{lemma}[Adjusted third order moments]\label{thm:mixGauss}
We have\beq \label{eqn:cross-moment} M_3= \sum_{j \in [r]}\rho_j w_j \cdot u_j \otimes u_j \otimes u_j,
\eeq where $\rho_j:=\Ebb[\nabla_{x'_j}^{(3)} g(x'_j)]$ and the expectation is over $x'_j=\inner{u_j,x}+\tilde{b}_j$. \end{lemma}

The proof follows from Stein's Identity. See Appendix~\ref{app:proofcentered} for details. 
Having the CP-form allows us to recover the component weight vectors through the tensor decomposition method. We present the result below.

\begin{theorem}[Recovery of mixture of GLMs] \label{thm:class_Gauss}
Assuming that the weight matrix $U\in \Rbb^{d\times r}$ is full column rank, $\rho_j,w_j \neq 0 ~~\forall ~ j \in [r]$, given $M_3$, we can recover the component weight vectors $u_j, j \in [r],$  up to scaling, using tensor method given in Algorithm~\ref{alg:robustpower} (in the Appendix).

\end{theorem}

The proof follows from Lemma~\ref{thm:mixGauss}. The computational complexity of tensor decomposition in this factor form is $O(nrdL)$, where $n$ is the number of samples and $L$ is the number of initialization.
Having recovered the normalized weight vectors, we can then estimate the scaling and the biases through expectation maximization or other methods. These are just $2r$ additional parameters, and thus, the majority of the parameters are estimated by the tensor method.

\begin{theorem}[Sample Complexity] \label{thm:sample-class-gauss}
Assume the conditions for Theorem~\ref{thm:class_Gauss} are met. Suppose the sample complexity
\begin{align*}
n =  \tilde{O}\left(\frac{d^3 r^4}{\epsilon^2 s_{\min}^2(U)}\right),
\end{align*}
then for each weight vector $u_j$, the estimate $\hat{u}_j$ from line~\ref{line:weight} Algorithm~\ref{algo:main} satisfies w.h.p
\begin{align*}
\Vert u_j-\hat{u}_j \Vert \leq \tilde{O}(\epsilon),~~ j \in [r].
\end{align*}
\end{theorem}
\paragraph{Proof outline: }From Lemma~\ref{thm:mixGauss}, we know that the exact cross-moment $\E[{y} \cdot \Sc_3(x)]$ has rank-one components as columns of matrix $U$; see Equation~\eqref{eqn:cross-moment} for the tensor decomposition form. Thus given the exact moment, the theorem is proved by applying the tensor decomposition guarantees in~\citet{JMLR:v15:anandkumar14b}.
In the noisy case where the moment is empirically formed by observed samples, we use the analysis and results of tensor power iteration in~\citet{anandkumar2014guaranteed}. They show that when the perturbation tensor is small, the tensor power iteration initialized by the SVD-based Procedure~\ref{algo:SVD init} in the Appendix recovers the rank-1 components up to some small error. The sample complexity is also proved by applying standard matrix concentration inequalities. In particular, we matricize the error tensor between exact moment and the empirical moment, and bound its norm with matrix Bernstein's inequality.

\paragraph{Remark :} We can also handle the case when the full column rank assumption on $U\in \Rbb^{d \times r}$ is violated under some additional constraints. In the overcomplete regime, we have the latent dimensionality exceeding the input dimensionality, i.e. $r> d$. The tensor method can still recover the weight vectors $u_j$, if we assume they are incoherent. A detailed analysis of overcomplete tensor decomposition is given in~\citep{anandkumar2014guaranteed}.

\paragraph{Remark :} If we assume the $u_j$ are normalized, the above approach suffices to completely learn the parameters $w_j$. This is because we obtain $w_j \rho_j$ and we have the knowledge of $\rho_j$, where the activation function and the input distributions are known. Otherwise, we need to perform EM to fully learn the weights. Note that initializing with our method results in performing EM in a low dimension instead of input dimension.  The reason is that the only unknown parameters are the scale and biases of the components. We initialize with the output of our method (Algorithm~\ref{algo:main}) and proceed with EM algorithm as proposed by~\citet{xu1995alternative}. For details see Appendix~\ref{sec:EM}.

\paragraph{Remark: }If $\rho_j=0$, which is the case for mixture of linear regression, we cannot recover the weight vectors from the tensor given in~\eqref{eq:simple_third}. In this case, we form a slightly different tensor to recover the weight vectors. We elaborate on this in the next section.

\paragraph{Remark: } Our results can be easily extended to multi-label and multi-class settings (one-versus-all strategy) as well as vector-valued regression problems.

\subsection{Learning Mixtures of Linear Regression}
We now consider mixtures of linear regressions:
\begin{align*}
\Ebb[y|x,h=e_j]= w_j \inner{u_j,x}+b_j,
\end{align*}
where $e_j \in \mathbb{R}^r$ denotes the $j$-th basis vector.

In this case higher order derivatives ($m \geq 2$) of the activation function vanish. Therefore, the cross-moment matrix and tensor defined in Section~\ref{sec:gausstoy} can not yield the parameters. For this setting, we form
 \begin{align}
 M_2&:=\Ebb[y^2 \cdot (x\otimes x-I)] 
 \nonumber \\
  M_3&:=\Ebb[y^3 \cdot x\otimes x \otimes x]-\sum_{j \in [d]} \!\Ebb[y^3 \cdot e_j \otimes x\otimes e_j]  \label{eq:tensor-reg}  \\
  &~-\!\!\sum_{j \in [d]} \!\Ebb[y^3 \cdot e_j \otimes e_j\otimes x] \!-\!\! \sum_{j \in [d]}\!\Ebb[y^3 \cdot x \otimes e_j \otimes e_j]. \nonumber
  \end{align}

\begin{lemma}[Adjusted third order moments]\label{thm:mixregGauss}
We have\beq M_3= \sum_{j \in [r]}\tilde{\rho}_j w_j \cdot u_j \otimes u_j \otimes u_j.
\eeq 
 \end{lemma}

The proof follows from Stein's Identity and it is provided in Appendix~\ref{app:proofcentered2}. Having the CP-form allows us to recover the component weight vectors through the tensor decomposition method. We present the result below.

\begin{theorem}[Recovery of linear regression mixtures] \label{thm:reg_Gauss}
Assuming that the weight matrix $U\in \Rbb^{d\times r}$ is full column rank, $\rho_j,w_j \neq 0 ~~\forall ~ j \in [r]$, given $M_3$ as in~\eqref{eq:tensor-reg}, we can recover the component weight vectors $u_j, j \in [r],$  up to scaling, using tensor method given in Algorithm~\ref{alg:robustpower} (in the Appendix).

\end{theorem}

The proof is similar to Lemma~\ref{thm:mixregGauss}.

\begin{theorem}[Sample Complexity] \label{thm:sample-reg-gauss}
Assume the conditions for Theorem~\ref{thm:reg_Gauss} are met. Suppose the sample complexity
\begin{align*}
n =  \tilde{O}\left(\frac{d^3 r^4}{\epsilon^2 s_{\min}^2(U)}\right),
\end{align*}
then for each weight vector $u_j$, the estimate $\hat{u}_j$ in line~\ref{line:weight} Algorithm~\ref{algo:main}  satisfies w.h.p
\begin{align*}
\Vert u_j-\hat{u}_j \Vert \leq \tilde{O}(\epsilon),~~ j \in [r].
\end{align*}
\end{theorem}
The proof follows the same approach as the one described for Theorem~\ref{thm:sample-class-gauss}.

\section{Learning GLM Mixtures under General Input Distribution} \label{sec:general_distr}

In the previous section, we established consistent estimation of the parameters of mixture of GLMs under Gaussian input. However, this assumption is limiting, since the input is usually far from Gaussian in any real scenario. We now extend the results in the previous section to any general (continuous) input.

 \subsection{Extensions of Stein's identity} \label{sec:stein}

The key ingredient that  enabled learning in the previous section is the   ability to  compute the expected derivatives of the output as a function of the input. Stein's identity shows that these derivatives can be obtained using  the cross-moments between the output and the score function of input. Is there a general unified framework where we can compute the expected derivatives under any general input distribution?

\citet{janzamin2014matrix} provide an affirmative answer. They show that by computing the cross-moment between the output and the (higher order) score functions of the input, we compute expected derivatives of any order. This key result allows us to extend the results in the previous section to any general input distribution.

\paragraph{Definition: Score function} The score of $x \in \mathbb{R}^d$ with pdf $p(x)$, denoted by $\Pc_1(x)$, is the random vector $\nabla_x \log p(x)$.~\citet{janzamin2014matrix},  define the $m^{\tha}$ order score function as
\beq \label{eqn:high-score}\Sc_m(x) := (-1)^m \dfrac{\nabla^{(m)} p(x)}{p(x)}.\eeq
They have also shown that score function can be equivalently derived using the recursive form
\begin{align}
\label{eqn:diffoperator_recursion_informal}
\Pc_m(x)& = - \Pc_{m-1}(x) \otimes \nabla_x \log p(x) - \nabla_x \Pc_{m-1}(x).
\end{align}

\begin{theorem}[Higher order derivatives~\citep{janzamin2014matrix}] \label{thm:steins_higher}
For random vector $x \in \R^{d}$, let $p(x)$ and $\Pc_m(x)$ respectively denote the pdf and the corresponding $m$-th order score function. Consider any continuously differentiable output-function $\Ebb[y|x]=g(x) :\mathbb{R}^{d} \rightarrow  \mathbb{R}$ satisfying some mild regularity conditions. Then we have
\[
\Ebb\left[ y \cdot \Pc_m(x) \right] =  \Ebb \left[ g(x) \cdot \Pc_m(x) \right] = \Ebb \left[ \nabla^{(m)}_x g(x) \right].
\]
\end{theorem}

For details, see~\citep{janzamin2014matrix}. In order to learn mixture of GLMs for general input distributions,  we utilize score function  $\Pc_m(\cdot)$ of order $m=3$.

\subsection{Moment forms}
We now consider the cross-moment $M_3:=\Ebb[y \cdot \Pc_3(x)]$, which is a third order tensor. By alluding to Theorem~\ref{thm:steins_higher}, we show that the moment $M_3$ has a CP decomposition where the components are the weight vectors $u_j$'s.

\begin{theorem}[Recovery of mixture of GLMs under general input]Given score function $\Pc_3(x)$ as in \eqref{eqn:high-score}, we have \begin{align*}
M_3:=\Ebb[y \cdot \Pc_{3}(x)]=\sum_{j \in [r]}\rho_j\cdot w_j \cdot u_j^{\otimes 3},
\end{align*}where $\rho_j:=\Ebb[\nabla_{x'_j}^{(3)} g(x'_j)]$ and the expectation is with respect to $x'_j:=\inner{u_j, x} + \tilde{b}_j$.

Assuming that matrix $U$ is full column rank and   $\rho_j,w_j > 0,~\forall j$, we can recover the weight vectors $u_j, j \in [r]$,  up to scaling, using tensor decomposition on $M_3$ given in Algorithm~\ref{alg:robustpower} (in the Appendix).
\end{theorem}

The proof follows from Theorem~\ref{thm:steins_higher}.
Thus, we have a guaranteed recovery of the weight vectors of mixture of GLMs under any general input distribution.

\paragraph{Remark: Sample complexity: } For general input sample complexity can be found in a similar approach to Gaussian case. The general form is  $n \geq  \tilde{O}\left(\Ebb\left[ \Vert H_3(x)H_3^\top(x)\Vert \right]\frac{d^{1.5} r^4}{\epsilon^2 s_{\min}^2(U)}\right)$. Here $H_3(x) \in \mathbb{R}^{d \times d^2}$ is the matricization of $\Pc_3(x)$. Theorem~\ref{thm:sample-class-gauss} follows from the fact that for Gaussian input $\Ebb\left[ \Vert H_3(x)H_3^\top(x)\Vert \right]=O(d^{1.5})$.

\paragraph{Remark: Score function estimation: } There are various efficient methods for estimating the score function. The framework of score matching is popular for parameter estimation  in probabilistic models~\citep{hyvarinen2005estimation, swersky2011autoencoders}, where the criterion is to fit parameters based on matching the data score function. For instance, \citet{swersky2011autoencoders} analyzes fitting the data to RBM (Restricted Boltzmann Machine) model.
Therefore, one option is to use this method for estimating $\Pc_1(x)$ and use the recursive form in~\eqref{eqn:diffoperator_recursion_informal} to estimate higher order score functions for the active layer.

\paragraph{Remark: Computational Complexity: } If we fit the input data into an RBM model, the computational complexity of our method, when performed in parallel, is $O(\log (\min (d, d_h)))$ with $O(r n L d d_h/\log(\min(d,d_h)))$ processors. Here $d_h$ is the number of neurons of the first layer of the RBM used for approximating the score function.

\subsection{Learning Mixtures of Linear Regression}
As discussed earlier, our framework can easily handle the case of mixtures of linear regression. Here, we describe it under general input distribution. Let,
\begin{align*}
\Ebb[y|x,h=e_j]= w_j \inner{u_j,x}+b_j,
\end{align*}
where $x$, $y$ respectively denote the input and output, and $h$ is the hidden variable that chooses the regression parameter $u_j$ from the set $\lbrace u_j \rbrace_{j \in [r]}$, $w_j= p(h=e_j)$ and $b_j$ is the bias.

\begin{theorem}[Recovery of linear regression mixtures under general input]\label{thm:regGauss}
Given score function $\Pc_3(x)$ as in Equation~\eqref{eqn:high-score}, we have \begin{align*}
M_3=\Ebb[y^3 \cdot \Pc_{3}(x)]=\sum_{j \in [r]} w_j \cdot u_j^{\otimes 3}.
\end{align*}

Assuming that matrix $U$ is full column rank,   $w_j \neq 0, ~~\forall j$ , we can recover the weight vectors $u_j, j \in [r]$,  up to scaling, using tensor decomposition on $M_3$ given in Algorithm~\ref{alg:robustpower} (in the Appendix).
\end{theorem}

For proof, see Appendix~\ref{app:proofcentered3}.
%


\paragraph{Remark: Sample complexity: } For general input sample complexity can be found in a similar approach to Gaussian case. The general form is  $n \geq  \tilde{O}\left(\Ebb\left[ \Vert H_3(x)H_3^\top(x)\Vert \right]\frac{d^{1.5} r^4}{\epsilon^2 s_{\min}^2(U)}\right)$. Here $H_3(x) \in \mathbb{R}^{d \times d^2}$ is the matricization of $\Pc_3(x)$. Theorem~\ref{thm:sample-reg-gauss} follows from the fact that for Gaussian input $\Ebb\left[ \Vert H_3(x)H_3^\top(x)\Vert \right]=O(d^{1.5})$.

\paragraph{Remark: }~\citet{tejasvi2013spectral}  consider learning a   mixture of  linear regression models,  using  tensor decomposition approach on the  higher order moments of the output $y$. They model the problem as an optimization on a third order tensor and prove that the optimal tensor would have the weight vectors as its rank-$1$ components. Minimizing an objective function over a tensor variable is expensive (in fact, quadratic for each variable~\citep{liu2009interior}, and their computational complexity scales as  $O(n d^{12})$. Hence their proposed method is not practical in large scale. Whereas, as discussed earlier our computational complexity is $O(n d^2)$. While we require the additional knowledge of the input distribution, in many scenarios, this is not a major limitation since there are large amounts of unlabeled samples which can be used for model estimation. Moreover, we can handle non-linear mixtures, while~\citet{tejasvi2013spectral} limit to linear ones.

\subsection{Extension to Mixture of GLMs with Nonlinear Transformations}

We have so far provided guarantees for learning mixture of GLMs. We now extend the results to cover non-linear models. We consider the class of mixture of GLMs with nonlinear transformations under the realizable setting as
\beq\label{eqn:latentsvm}
\Ebb[ y| x,h ]=g\left( \inner{ Uh, \phi(x)}+\tilde{b} \right),
\eeq  where $\phi(x)$ represents the nonlinear mapping of $x$. Assuming that $\phi(\cdot)$ is known, we propose simple ideas to extend our previous results to the setting in \eqref{eqn:latentsvm}.

The key idea is to compute the score function $\Pc_m(\phi(x))$ corresponding to $\phi(x)$ rather than the input $x$. There is a simple relationship between the scores.
The connection can be made from the probability density of the transformed variable as follows.
Let $t=\phi(x)$, $D_t(i,j):=\left[ \frac{\partial x_i}{\partial t_j} \right]$. We have
\begin{align}
\label{eqn:score_nl}
p_{\phi(x)}(t_1, \cdots, t_p)&=p_x(\phi_1^{-1}(t), \cdots, \phi_p^{-1}(t)) \vert \det(D_t) \vert, \\
\mathcal{S}_m(t)&=(-1)^m \frac{\nabla^{(m)}_t p_{\phi(x)}(t)}{p_{\phi(x)}(t)}. \nonumber
\end{align}

\begin{theorem}[Recovery of mixture of GLMs with nonlinear transformations under general input]Given score function $\Pc_3(\phi(x))$ as in Equation~\eqref{eqn:score_nl}, we have \begin{align*}
M_3:=\Ebb[y \cdot \Pc_{3}(\phi(x))]=\sum_{j \in [r]}\rho_j\cdot w_j \cdot u_j^{\otimes 3},
\end{align*}where $\rho_j:=\Ebb[\nabla_{z_j}^{(3)} g(z_j)]$ and the expectation is with respect to $z_j:=\inner{u_j, \phi(x)} + \tilde{b}_j$.

Assuming that matrix $U$ is full column rank, $w_j,\rho_j \neq 0, ~~\forall j$, we can recover the weight vectors $u_j, j \in [r]$,  up to scaling, using tensor decomposition on $M_3$ given in Algorithm~\ref{alg:robustpower} (in the Appendix).
\end{theorem}

We therefore have a guaranteed recovery of the parameters of mixture of GLMs with nonlinear transformations under the realizable setting, given score function $\Pc_3(\phi(x))$.


\section{Related Works}


\textbf{Mixture of Experts/ Regression Mixtures: }The mixture of experts model was introduced as an efficient probabilistic ``divide'' and ``conquer'' paradigm in~\citep{jordan1994hierarchical}. Since then, it has been considered in a number of works, e.g.~\citep{xu1995alternative,svensen2003bayesian}. Learning is carried out usually through   EM~\citep{jordan1994hierarchical,xu1995alternative} or variational approaches~\citep{svensen2003bayesian}, but the methods have  no guarantees. Works with guaranteed learning of associative mixture models are fewer.~\citet{tejasvi2013spectral}  consider learning a   mixture of  linear regression models,  using  tensor decomposition approach on the  higher order moments of  of the label $y$.~\citet{yi2013alternating} also  consider mixed linear regression problem  with two components and  provide consistency guarantees in the noiseless setting for an alternating minimization method.~\citet{chen2013convex} provide an alternative convex method for the same setting under noise and established near optimal sample complexity. However, all these guaranteed methods are restricted to mixture of linear regressions and do not extend to  non-linear models.


\textbf{Learning  mixture of GLMs: }For the mixture of generalized linear models (GLM),~\citet{li1992principal} and~\citet{SunIM13} present methods for learning the subspace of the weight vectors of the component GLMs, assuming that the input is white  Gaussian distribution.~\citet{li1992principal} propose the so-called  principal Hessian directions (PHd), where the eigenvectors of the second-order moment matrix $\Ebb[y\cdot x\otimes x]$ are used to learn the desired subspace (the notation $\otimes$ represents tensor (outer) product). However,  the PHd method fails  when the output $y$ is a symmetric function of the input $x$, since the moment matrix vanishes in this case.~\citet{SunIM13} overcome this drawback through their clever ``mirroring'' trick   which transforms the output $y$ to $r(y)$ such that the resulting second order moment $\Ebb[r(y) \cdot x \otimes x]$ matrix does not vanish.

Our work has some key differences: the works in~\citep{li1992principal,SunIM13} assume Gaussian input $x$, while we allow for any probabilistic model (with continuous density function).  
 Another important difference between~\citep{li1992principal,SunIM13} and our work, is that we use tensor-based learning techniques, while~\citep{li1992principal,SunIM13} only operate on matrices. Operating on tensors allows us to learn the individual weight vectors (up to scaling) of the mixture components, while~\citep{li1992principal,SunIM13} only learn the subspace of the weight vectors. 


\textbf{Spectral/Moment based methods for discriminative learning: }
\citet{karampatziakis2014discriminative} obtain discriminative features via generalized eigenvectors. They consider the tensor $\Ebb[y \otimes x \otimes x]$ and then treat $\Ebb[x\otimes x| y=i]$ as  the signal for class $i$ and $\Ebb[x \otimes x|y=j]$ as the noise due to class $j$. They contrast their method against classical discriminative procedures such as Fisher LDA and show good performance on many real datasets. However, their method has some drawbacks: they cannot handle continuous $y$, and also   when $y$ has a large number of classes $m$ and $x\in \Rbb^d$ has high dimensionality, the method is not scalable since it requires $m^2$ eigen-decompositions of $d \times d $ matrices. Another line of moment based methods are the so-called {\em sliced inverse regression} (SIR)~\citep{li1991sliced}, where input  $x$ is regressed against output $y$.  These methods project the input to a lower dimension subspace that preserves the required information.~\citet{li1991sliced} consider  top eigen components of the moment $\Ebb[\Ebb[x|y] \Ebb[x|y]^\top]$ for dimensionality reduction. 
\section{Conclusion}
In this paper, we propose a tensor method for efficient   learning of associative mixtures.   In addition to employing the learnt weight vectors in the mixture of GLMs model for prediction, we can employ them in a number of alternative ways in practice. For instance, we can utilize the output of the tensor-based methods as initializers for likelihood based techniques such as expectation maximization. Since these objective functions are non-convex, in general, they can get stuck in bad local optima. Initializing with the tensor methods can lead to convergence to better local optima. Moreover, we can employ the learnt weight vectors to construct discriminative features and train a different classifier using them. Thus, our method yields discriminative information which is useful in myriad ways.

There are many future directions to consider.
We assume that the choice variable for selecting the mixture components is independent of the input. This is also the assumption in a number of other works for learning regression/classifier mixtures~\citep{SunIM13,tejasvi2013spectral}. In the general mixture of experts framework, the choice variable is known as the gating variable, and it selects the classifier based on the input.  Considering this scenario is of interest. Moreover, we have considered continuous input distributions,  extending this framework to discrete input is of interest.

%


\subsection*{Acknowledgment}
H. Sedghi is supported by NSF Career award FG15890. M. Janzamin is supported by NSF BIGDATA award FG16455. A. Anandkumar is supported in part by Microsoft Faculty Fellowship, NSF Career award CCF-$1254106$,   and ONR Award N00014-14-1-0665.


\appendix

\section{Proofs}
\subsection{Proof of Lemma~\ref{thm:mixGauss}} \label{app:proofcentered}
{\bf Notation: Tensor as multilinear forms:} We view a tensor $T \in \Rbb^{d \times d \times d}$ as a multilinear form. 
Consider matrices $M_r \in \R^{d\times d_r}, r \in \{1,2,3\}$. Then tensor $T(M_1,M_2,M_3) \in \R^{d_1}\otimes \R^{d_2}\otimes \R^{d_3}$ is defined as
\begin{align} \label{eqn:multilinear form def}
T(M_1,M_2,M_3)_{i_1,i_2,i_3} := \sum_{j_1, j_2,j_3\in[d]} T_{j_1,j_2,j_3} \cdot M_1(j_1, i_1) \cdot M_2(j_2, i_2) \cdot M_3(j_3, i_3).
\end{align}
In particular, for vectors $u,v,w \in \R^d$, we have\,\footnote{Compare with the matrix case where for $M \in \R^{d \times d}$, we have $ M(I,u) = Mu := \sum_{j \in [d]} u_j M(:,j) \in \R^d$.}
\begin{equation} \label{eqn:rank-1 update}
 T(I,v,w) = \sum_{j,l \in [d]} v_j w_l T(:,j,l) \ \in \R^d,
\end{equation}
which is a multilinear combination of the tensor mode-$1$ fibers.
Similarly $T(u,v,w) \in \R$ is a multilinear combination of the tensor entries,  and $T(I, I, w) \in \R^{d \times d}$ is a linear combination of the tensor slices.

Now, let us proceed with the proof.

\bprf Let $x':= \inner{u,x}+b$.  Define $l(x) := y \cdot x \otimes x$. We have
\begin{align*}\Ebb[y\cdot x^{\otimes 3}]&= \Ebb[l(x) \otimes x]
=  \Ebb[\nabla_x l(x)],
\end{align*} by applying Stein's lemma. We now simplify the gradient of $l(x)$.
\begin{align}  {\mathbb{E}\left[ \nabla_x l(x)  \right]}=\Ebb[ y \cdot\nabla_x (x\otimes x)] +\Ebb[(\nabla_{x'} g(x')) (x\otimes x\otimes u)].\label{eqn:twoterms}
\end{align}

 We now analyze the first term.  We have
\bcase{\nabla_x (x\otimes x)_{i_1,i_2,j}= \frac{\partial x_{i_1} x_{i_2}}{\partial x_j}=} x_{i_2}, & $i_1=j$,\nn \\ x_{i_1}, &  $i_2=j$,  \\ 2x_{j}, & $i_1=i_2=j$,\nn \\ 0, & o.w.\nn \ecase
This can be written succinctly as
\begin{align*} \nabla_x  (x\otimes x) =  \sum_i e_i \otimes x\otimes e_i + \sum_i x \otimes e_i \otimes e_i + \sum_i 2 x_i (e_i \otimes e_i \otimes e_i) \end{align*} and therefore, the expectation for the first term  in \eqref{eqn:twoterms} is given by  \[ \Ebb[ y\cdot\nabla_x(x\otimes x)]=
\sum_i \left(\Ebb[y \cdot e_i \otimes x\otimes e_i] + \Ebb[y \cdot x \otimes e_i \otimes e_i]+{2} \Ebb[y \cdot x_i \cdot e_i\otimes e_i \otimes e_i]\right).\]
Now for the second term in \eqref{eqn:twoterms}, let $f(x):= \nabla_{x'} g(x') \cdot  x\otimes u$. The transposition of the second term in \eqref{eqn:twoterms} is given by \begin{align*} \Ebb[ \left(\nabla_{x'} g(x') \cdot  x\otimes u\right)\otimes x]&= \Ebb[  f(x)\otimes x]\\ &= \Ebb[ \nabla_x f(x)], \end{align*}
where we have swapped modes $2$ and $3$ in $\Ebb[(\nabla_{x'} g(x')) (x\otimes x\otimes u)]$ to obtain the above. We will compute $\nabla_x f(x)$ and then switch the tensor modes again to obtain the final result.
We have
\begin{align}\nn\nabla_x f(x) &= \nabla_x \left(\nabla_{x'} g(x') x \otimes u\right)\\ \label{eqn:twotermsf}& = (\nabla^{(2)}_{x'} g(x'))\cdot  x \otimes u \otimes u + (\nabla_{x'} g(x'))\cdot \nabla_x (x \otimes u),\end{align}
The first term is given by \[ \Ebb\left[(\nabla^{(2)}_{x'} g(x'))\cdot  x \otimes u \otimes u\right] = \Ebb\left[(\nabla^{(3)}_{x'} g(x'))\cdot  u \otimes u \otimes u\right] \]
So the second term in \eqref{eqn:twotermsf} is given by
\[ \sum_{i} (\nabla_{x'} g(x'))\cdot (e_i \otimes u \otimes e_i).\]
Note that
\[  \Ebb\left[(\nabla_{x'} g(x'))\cdot (e_i \otimes u \otimes e_i)\right]= \Ebb\left[ e_i \otimes \nabla_x g(\inner{x,u}) \otimes e_i\right] = \Ebb\left[g(x') \cdot (e_i \otimes x \otimes e_i) \right],\] since if we apply Stein's left to right-hand side, we obtain the left hand side of the equation.  Swapping the modes $2$ and $3$ above, we obtain the result by substituting in \eqref{eqn:twoterms}. 

We need to mention that, Lemma~\ref{thm:mixGauss}  can be directly proved by Theorem~\ref{thm:steins_higher} as specific form of score function for Gaussian input. Here,  we have provided step by step first principles proof of the lemma for easy understanding.
\eprf

\subsection{Proof of Lemma~\ref{thm:mixregGauss}} \label{app:proofcentered2}
By replacing $y$ by $y^3$ in Proof of Lemma~\ref{thm:mixGauss} (Appendix~\ref{app:proofcentered}), we have that
\begin{align*}
M_3&=\Ebb_x \left[\nabla_x^3 (y^3)\right]=\Ebb_x\left[\nabla_x^3\Ebb_h\left[ y^3|h=e_j\right]\right]\\&=\Ebb_x\left[\nabla_x^3 \left(\sum_{j \in [r]} \left(w_j \inner{u_j,x}+b_j\right)^3\right)\right]
= \sum_{j \in [r]}\rho_j w_j \cdot u_j \otimes u_j \otimes u_j.
\end{align*}

Note that the third equation results from the fact that for each sample only one of the $u_j, j \in [r]$ is chosen by $h$ and  no other terms are present. Therefore,the expression has no cross terms.

\subsection{Proof of Theorem~\ref{thm:regGauss}} \label{app:proofcentered3}
\begin{proof}
\begin{align*}
M_3&=\Ebb_x[y^3 \cdot \Pc_{3}(x)]
=\Ebb_h\left[\Ebb_x[y^3 \cdot \Pc_{3}(x)|h=e_j]\right]\\&
=\Ebb_x\left[\Ebb_h[y^3 \cdot \Pc_{3}(x)|h=e_j]\right]
=\Ebb_x\left[\sum_{j \in [r]} \left(w_j \inner{u_j,x}+b_j \right)^3 \right]\\&=\Ebb_x\left[\nabla^3_{x} \left[\sum_{j \in [r]} \left(w_i \inner{u_j,x}+b_j\right)^3 \right]\right]
=\sum_{j \in [r]} w_j \cdot u_j^{\otimes 3}.
\end{align*}
Note that the fourth equation results from the fact that for each sample only one of the $u_j, j \in [r]$ is chosen by $h$ and  no other terms are present. Therefore,the expression has no cross terms.
\end{proof}

\section{Tensor Decomposition Method} \label{sec:tensor_decomposition}
We now recap the tensor decomposition method~\cite{anandkumar2014guaranteed} to obtain the rank-$1$ components of a given tensor.  This is given in Algorithm~\ref{alg:robustpower}. 
Let $\widehat{M}_3$ denote the empirical moment tensor input to the algorithm.

Since in our case modes are the same, the asymmetric power updates in~\citep{anandkumar2014guaranteed} are simplified to one update. These can be considered as rank-1 form of the standard alternating least squares (ALS) method. If we assume the weight matrix $U$ (i.e. the tensor components) has incoherent columns, then we can directly perform tensor power method on the input tensor $\widehat{M}_3$ to find the components. Otherwise, we need to whiten the tensor first. We take a random slice of the empirical estimate of $\widehat{M}_3$ and use it to find the whitening matrix\footnote{If $\Ebb[y|x]$ is a symmetric function of $x$, then the second moment $M_2$ is zero. Therefore, we cannot use it for whitening. Instead, we use random slices of the third moment $M_3$  for whitening.}. Let $\widehat{V}$ be the average of the random slices. The whitening matrix $\widehat{W}$ can be found by using a rank-$r$ SVD on $\widehat{V}$ as shown in Procedure~\ref{algo:whitening}.

Since the tensor decomposition problem is non-convex, it requires good initialization. We use the initialization algorithm from~\citep{anandkumar2014guaranteed} as shown in Procedure~\ref{algo:SVD init}.
The initialization for different runs of tensor power iteration is performed by the SVD-based technique proposed in Procedure~\ref{algo:SVD init}. This helps to initialize non-convex power iteration with good initialization vectors when we have large enough number of initializations. Then, the clustering algorithm is applied where its purpose is to identify which initializations are successful in recovering the true rank-1 components of the tensor.



\floatname{algorithm}{Procedure}
\setcounter{algorithm}{1}
\begin{algorithm}[t]
\caption{Whitening}
\label{algo:whitening}
\begin{algorithmic}[1]
\renewcommand{\algorithmicrequire}{\textbf{input}}
\renewcommand{\algorithmicensure}{\textbf{output}}
\REQUIRE Tensor $T \in \Rbb^{d\times d \times d}$.
\STATE Draw a random standard Gaussian vector $\theta \sim \mathcal{N}(0,I_{d}).$
\STATE Compute  $\widehat{V}=T(I,I,\theta) \in \R^{d \times d}$.
\STATE Compute the rank-$r$ SVD $\widehat{V}=\tilde{U}\Diag(\tilde{\lambda})\tilde{U}^\top$.
\STATE Compute the whitening matrix $\widehat{W}=\tilde{U}\Diag(\tilde{\lambda}^{-1/2})$.
\RETURN $T\left(\widehat{W},\widehat{W},\widehat{W}\right)$.
\end{algorithmic}
\end{algorithm}

\floatname{algorithm}{Procedure}
\begin{algorithm}[t]
\caption{SVD-based initialization when $r = O(d)$ ~\citep{anandkumar2014guaranteed}}
\label{algo:SVD init}
\begin{algorithmic}[1]
\renewcommand{\algorithmicrequire}{\textbf{input}}
\renewcommand{\algorithmicensure}{\textbf{output}}
\REQUIRE Tensor $T \in \Rbb^{r \times r \times r}$.
\STATE Draw a random standard Gaussian vector $\theta \sim \mathcal{N}(0,I_{r}).$
\STATE Compute $u_1$ as the top left and right singular vector of  $T(I,I,\theta) \in \R^{r \times r}$.
\STATE $\ha_0 \leftarrow u_1$.
\RETURN $ \ha_0 $.
\end{algorithmic}
\end{algorithm}

%
%
%
%
%
%
%
%
%
%
%
%
%

\floatname{algorithm}{Algorithm}
\begin{algorithm}[h]
\caption{Robust tensor power method~\citep{anandkumar2014guaranteed}}
\label{alg:robustpower}
\begin{algorithmic}[1]
\renewcommand{\algorithmicrequire}{\textbf{input}}
\renewcommand{\algorithmicensure}{\textbf{output}}
\REQUIRE symmetric tensor ${T} \in \R^{d \times d \times d}$, number of
iterations $N$, number of initializations $L$, parameter $\nu$.

\ENSURE the estimated eigenvector/eigenvalue pair. 

\STATE Whiten ${T}$ using the whitening method n Procedure~\ref{algo:whitening}.
\FOR{$\tau = 1$ to $L$}

\STATE Initialize $\ha_0^{(\tau)}$ with SVD-based method in Procedure~\ref{algo:SVD init}.

\FOR{$t = 1$ to $N$}

\STATE Compute power iteration update
\begin{eqnarray}
\ha_{t}^{(\tau)} & := &
\frac{{T}(I, \ha_{t-1}^{(\tau)}, \ha_{t-1}^{(\tau)})}
{\|{T}(I, \ha_{t-1}^{(\tau)}, \ha_{t-1}^{(\tau)})\|}
\label{eq:power-update}
\end{eqnarray}

\ENDFOR

\ENDFOR


\STATE $S := \left\{ a_\tau^{(N+1)}:\tau\in [L]\right\}$

\WHILE{$S$ is not empty} 
\STATE Choose $a \in S$ which maximizes $|T(a,a,a)|$.
\STATE Do $N$ more iterations of \eqref{eq:power-update} starting from $a$.
\STATE {\bf Output} the result of iterations denoted by $\ha$.
\STATE Remove all the $a \in S$ with $|\langle a, \ha\rangle| > \nu/2$.
\ENDWHILE
%
%

\end{algorithmic}
\end{algorithm} 

\section{Expectation Maximization for Learning Un-normalized Weights} \label{sec:EM}

If we assume the weight vectors are normalized, our proposed algorithm suffices to completely learn the parameters $w_i$. Otherwise, we need to perform EM to fully learn the weights. Note that initializing with our method results in performing EM in a lower dimension than the input dimension. In addition, we can also remove the independence of selection parameter from input features when doing EM.  We initialize with the output of our method (Algorithm~\ref{algo:main}) and proceed with EM algorithm as proposed by~\citet{xu1995alternative}, Section 3. Below we repeat the procedure in our notation for completeness.

Consider the gating network
\begin{align*}
g_j(x,\nu)&=\frac{w_j p(x|\nu_j)}{\sum_i w_i p(x|\nu_i)}, \quad \sum_i w_i=1, \quad w_i \geq 0, \\
p(x|\nu_j)&=a_j(\nu_j)^{-1}b_j(x)\exp\lbrace c_j(\nu_j)^\top t_j(x) \rbrace,
\end{align*}
where $\nu = \lbrace w_j, \nu_j, j=1, \cdots, r \rbrace$, and the $p(x| \nu_j)$'s are density functions from the exponential family. 

In the above equation, $g_j(x,\nu)$ is actually the posterior probability $p(j|x)$ that $x$ is assigned to the partition corresponding to the $j-$th expert net. From Bayes' rule:
\begin{align*}
g_j(x, \nu)=p(j|x)=\frac{w_j p(x| \nu_j)}{p(x,\nu)}, \quad p(x,\nu)=\sum_i w_i p(x|\nu_i). 
\end{align*}
Hence,
\begin{align*}
p(y|x,\Theta) = \sum_j \frac{w_j p(x| \nu_j)}{p(x,\nu)} p(y|x,u_j),
\end{align*}
where $\Theta$ includes $u_j, j=1, \cdots, r$ and $\nu$.
Let
\begin{align*}
Q^g(\nu) &= \underset{t}{\sum}\underset{j}\sum f_j^{(k)}(y^{(t)}|x^{(t)}) \ln g_j^{(k)} (x^{(t)},\nu^{(t)}),\\
Q_j^g(\nu_j) &= \underset{t}{\sum} f_j^{(k)}(y^{(t)}|x^{(t)}) \ln p(x^{(t)} | \nu_j),\quad j \in [r]\\
Q_j^e(\theta_j) &=\underset{t}{\sum} f_j^{(k)}(y^{(t)}|x^{(t)}) \ln p(y^{(t)}|x^{(t)},\theta_j), \quad j \in [r]\\
Q^{w} &= \underset{t}{\sum}\underset{j}\sum f_j^{(k)}(y^{(t)}|x^{(t)}) \ln w_j, \quad \text{with} ~~ w=\lbrace w_1, \dotsc, w_r \rbrace
\end{align*}
The EM algorithm is as follows:
\begin{enumerate}
\item E-step. Compute
\begin{align*}
f_j^{(k)}(y^{(t)}|x^{(t)})=\frac{w_j^{(k)}p(x^{(t)}|\nu_j^{(k)}) p(y^{(t)}|x^{(t)},u_j^{(k)})}{\sum_i w_i^{(k)}p(x^{(t)}|\nu_i^{(k)})p(y^{(t)}|x^{(t)},u_i^{(k)})}. 
\end{align*}
\item M-Step Find a new estimate for $j=1, \cdots, r$
\begin{align*}
u_j^{(k+1)}&=\underset {u_j}{\arg \max}~ Q_j^e(u_j), \quad \nu_j^{(k+1)}=\underset {\nu_j}{\arg \max}~ Q_j^g(\nu_j), \\
w^{(k+1)}&=\underset {w}{\arg \max}~ Q^w,~~ \text {s.t.}~~ \sum_i w_i=1.
\end{align*}
\end{enumerate}




\end{document}